# A global AI community requires language-diverse publishing

**Haley Lepp and Parth Sarin, Stanford University**

Who produces a global AI? Who researches, designs, builds, trains, evaluates, and sells the models that are being deployed across the planet? An internationally diverse network of actors contributes to the technologies known as AI, including the "ghost" laborers who extract the material resources for energy and hardware [7], the barely-wage earners who label and label and label [5]; and the non-wage earning "users" whose labor has been classified by AI companies not as creation, but consumption [10]. In the spotlight are the AI researchers. Typically highly-paid corporate employees or scholars with high-earning potential, AI researchers have an out-sized voice in determining the trajectory of this technology, a voice which they project through publications and conference proceedings. Over and over again, this community has been interrogated for its lack of diversity in race, gender, country, and age. As a result, affinity groups have proliferated; conferences have adopted anti-bias policies, and volunteers have worked tirelessly to expand training opportunities for groups who have historically faced discrimination in computing communities. Yet a major component of diversity remains unaddressed, one which we argue helps uphold extractive and unbalanced nature of this powerful industry. To be an AI researcher in this global community, one must write and speak in English.

The top 100 ranked computer science proceedings and journals are published in English [1]. Despite their locations all over the world, in many countries in which English is not a primary language spoken, conference proceedings are officially held English. Even scholars who invest considerable time and expense into learning to produce academic English may face rejection in peer review. In fact, mining the historical reviews from the past six years of ICLR proceedings demonstrates that thousands of reviews over the years critique the language of authors either explicitly ("The paper is full of English mistakes.") or implicitly ("There are numerous grammatical errors and poorly-phrased sentences.") [12]. The ramifications of this monolingual research industry are wide: from heavily uneven linguistic output and poor attention to issues in so-called "low-resource" languages [4, 11], to limitations in global education and hiring [8], to a de facto tax for scholars who must pay for pre-submission copy-editing [2, 3].

What does a researcher do when they believe their work may be rejected based on reviewer perceptions of academic English? First, some will choose not submit work and simply not participate in the research community. Second, some will submit their work, but pay someone to translate or "professionalize" their writing. This process is expensive and requires extra time, which in such a fast-moving field puts such scholars at a competitive disadvantage.

As such, the increasing availability of automatic writing-assistance and translation software has been celebrated as a boon for inclusion, allowing scholars who might otherwise be excluded from monolingual publishing to instantly translate their writing prior to submission for peer review. Many scholars use ChatGPT to "fix" their writing for publication. However, we regard this phenomenon not as a solution for inclusion, but a symptom of linguistic exclusion. Authors should have the right to describe their findings in the language of their choice, without the mediation of translation. In preliminary interviews for a future study, multilingual ICLR scholars indicated that they feel that they have different personalities when they express themselves in one language over another. Translation is also never one-to-one, so by only publishing in one language, the community loses out on the vast diversity of other ways of knowing which might be represented. Furthermore, producing research in a single language also alienates readers in other languages. Widespread translation to English will drive our field away from, instead of toward, an inclusive global community.

Without intervention, two possibilities loom on the horizon. The AI publishing community will continue to exclude the majority of the world's language groups and therefore people, and only the limited people in the world who receive long-term English training will participate. Alternatively, current publishing infrastructure will demand the increased assimilation to English-only education and publishing. Computing training will be paired with English-language training, pushing English around the globe as a requirement for high-paid employment and intellectual contribution to technology research. Neither of these are tolerable futures for a truly global research community. So, where can we go from here?

First, conferences should be administered in the language(s) of the country in which they are held. Organizers should hire translation services, and encourage scholars who speak English and other languages to present in their other languages, un-hiding the linguistic diversity of the existing computing community.

Second, we should include explicit instructions to peer reviewers to not adjudicate the language "appropriateness" of the papers they review [6]; instead, editors should explore other options for publishing multilingual scholarship, such as offering scholars the opportunity to share their work in multiple languages. English-speakers should share any cost burden of translation, instead of placing it entirely on those who do not speak English. Publications and conferences must set aside funds for translation services- not just into English, but from English. Publication infrastructure, such as OpenReview, should also include space for submissions of translations.

Finally, it is imperative that the burden of change not be upon people who are linguistically marginalized in publishing: everyone in this global community should be working toward learning to tolerate and embrace linguistic diversity. Graduate students should be encouraged, if not required, to take language courses [9]. If we truly aspire to have a global AI community, we must challenge the hegemony of English in computing.

# 一个全球性的 AI 社群，需要语言多元化的出版业

Haley Lepp 和 Parth Sarin，斯坦福大学
房天语（Tianyu M. Fang）翻译

谁在推动全球的人工智能（AI）发展？谁在研究、设计、建造、训练、评估并销售正在全球各地部署的模型？我们所称为 AI 的技术依赖于一个国际化、多元化的参与者网络。这其中包括为能源和硬件提取物质资源的“幽灵”劳工[7]；依靠不断数据标注工作获得微薄薪水的工人[5]；以及那些劳动被 AI 企业视为“消费”而非“创造”的、不领取工资的“用户”[10]。然而，聚光灯更多地集中在 AI 研究人员身上。他们通常是高薪的企业员工，或是具备高收入潜力的学者，他们对决定技术发展方向有着重要的发言权，并通过期刊和会议论文集表达他们的观点。这个群体因其在种族、性别、国籍和年龄方面缺乏多样性而屡遭质疑。为此，各类同背景的组织（affinity groups）不断涌现；会议也采取了反偏见政策，义工们不懈努力，为计算机领域中历来受歧视的群体提供更多培训机会。然而，多样性的一个重要组成部分依然未得到足够的重视，而我们认为，这一要素维持了这个强大行业中的压迫与不平等。要成为这个全球社区中的 AI 研究人员，必须用英语进行写作和交流。

全球前 100 的计算机科学领域会议论文集和期刊全部用英语发表[1]。尽管这些会议在世界各地举行，即使在非英语母语国家，会议的正式语言依然是英语。即便是花费大量时间和金钱学习学术英语写作的学者，依然有可能在同行评审中被拒绝。事实上，对过去六年的 ICLR 会议评审记录的分析表明，多年来有数千篇评审意见明确地（如“该论文充满了英语错误”）或隐晦地（如“文中有多处语法错误和措辞不当的句子”）批评了作者的语言问题[12]。这种单一语言的研究行业产生了广泛的影响，包括语言输出的严重不均衡、对所谓“低资源”语言问题的忽视[4,11]、全球教育与招聘的限制[8]、以及学者们在发表论文前必须支付的“审稿税”[2,3]。

当研究人员认为其工作可能因评审人对其学术英语水平的看法而被拒绝时，他们会做些什么？首先，部分研究人员会选择不提交稿件、不参与研究社群的活动。其次，一些研究人员会提交稿件，但同时花钱请别人翻译或“润色”他们的写作。这个过程不仅昂贵，还需要额外的时间，使非母语学者在这个快速发展的领域中处于竞争劣势。
因此，随着越来越多的机器辅助写作和翻译软件被视为多元化的解决方案，那些被英语出版业排除在外的学者可以快速翻译他们的文章以提交同行评审。许多学者使用 ChatGPT“修补”他们的文章以便发表。然而，我们不认为这一现象是包容的解决方案，而是语言排斥的另一种表现。作者应该有权用自己选择的语言来描述研究成果，而不依赖翻译作为中介。在未来一项研究的初步访谈中，多语的 ICLR 学者表示，他们在不同语言的表达中展现出不同的个性。翻译从来都不是一对一的；因此，仅用一种语言发表研究，社群则会失去其他语言中所包含的多样化知识。此外，仅用一种语言进行研究也会疏远其他语言的读者。广泛的英语翻译将使我们的领域远离包容的全球社群，而不是向其靠近。

如果不采取干预措施，未来有两种可能性。一是 AI 出版社群将继续排除世界上大多数语言群体和人口，只有接受过长期英语训练的少数人才能参与其中。二是当前的出版基础设施将迫使人们更加适应仅以英语为主的教育和出版业。计算机培训将与英语培训相结合，导致英语能力成为全球高薪工作和对技术研究作出贡献的前提条件。这两种未来都不是一个真正全球化研究社区所能接受的。那么，我们该何去何从？

首先，会议应以举办国家的语言进行。组织者应聘请翻译服务，并鼓励既会说英语又会说其他语言的学者以本地语言进行演讲，还原现有计算机领域中的语言多样性。

其次，我们应为同行评审员提供明确指示，要求他们不对审阅论文语言的“适当性”进行评判[6]。相反，编辑应探索发表多语种学术作品的可能，例如为学者提供用多种语言发表学术文章的机会。英语使用者应分担翻译费用，而不是将全部负担加在不会说英语的人身上。出版物和会议必须预留资金用于翻译服务，不仅是将其他语言翻译成英语，还包括从英语翻译至其他语言。出版基础设施如 OpenReview，也应提供提交译文的选项。

最后，我们必须强调，改变现状的责任不应由出版中处于语言边缘化的人群来承担。全球社区中的每个人都应努力学习如何宽容对待并接受语言多样性。我们应鼓励甚至要求研究生选修语言课程[9]。如果我们真正渴望拥有一个全球性的 AI 社区，就必须挑战英语在计算机领域中的霸权地位。

# Tekotevẽ oñehendu opaite ñe'ẽ ñane ñomongeta IA rehegua

**Haley Lepp and Parth Sarin**

**Michael Hardy ombohasa ko kuatiahaipyre guaraníme**

Mávapa ojapo peteĩ IA mayma tetãyguápe? Mávapa oinvetiga, ohesa'ỹijóvo, ojapo, omopu'ã, ombokatupyry, ojesareko porã, ha ovende umi mba'e oñembohapéva hína opiate ñande yvy'ári rehegua? Heta oi umi omba'apóva opaichagua oĩva ko múndo tuichakuére ojapo ikatu haguãicha ojejapo disponible guarãumi tecnología ojekuaáva IA ramo, oimehápe umi mba'apohára "ojehecha'ỹvanguéra " kañymby oguenohêva ha'eva mba'apohára omba'apo mbaretéva ojepaga vaíva ha heta omba'apo pytu'u'ỹre.

Upéi oĩ tapicha oiporúva IA ojepaga'ỹre hese, ha'eháicha rehechávo video térã reiporúramo iselulárpe, ajépa. Umi ñahenduvéva ha'e umi ohesa'ỹijóvova and papapykuaahára IA rehegua. Jepive ojepaga porã chupekuéra ha oguereko tuicha ñe'ẽ mba'éichapa oñemoakãrapu'ã IA. Ohai kuatia ha oho aty guasúpe oikuaauka haguã hemiandu. Jeyjey, umikuéra oñe'ẽ mba'éichapa IA umi ohesa'ỹijóvova and papapykuaahára kuéra ndorekóimí tapicha iñambuéva - opaichagua hapichaichagua, oñe'ẽmbueicha, kuimba'e ha kuña mokoĩveva katu, tetã, ha arykuéra-icha.

Jeyjey, umikuéra oñe'ẽ mba'éichapa IA ojapóva kuéra ndorekóimíkena tapicha iñambuéva - opaichagua hapichaichagua, oñe'ẽmbueicha, kuimba'e ha kuña katu, tetã ha arykuéra-icha. Upéicha, oĩ tapicha oñepyrũva oipytyvõmana umi ojehejávape aty'ípe. Umi ojapóva aty guasu ojapo hekopeteporã opavavémipe ĝuarã, ha oĩete umi oipytyvõva hetave tapichápe oikuaa haĝua IA rehegua, ko'ýte umi ojetratava'ekue hekope'ỹ mboyve. Ha katu ofaltagueteri peteĩ mba'e: reñe'ẽva'erã ingléite-pe, ha'e haĝua peteĩ IA ojapóva-pe. Roimo'ã péva hekope'ỹha umi mba'e ha omantene umi mba'e noñemyatyrõi.

Umi porãveva arandukami ha aty guasúpe IA rehegua oĩmbaite inglé-pe, jepémo ke oiko oparupiete ko yvy ape ári, ajépa. Péa ikatu hasy tapichakuéra ndoingleñe'ẽporãivanguéra guarã. Oñeha'ãmbavéramo jepe, ikatu gueteri oñemboyke hembiapokuéra ojavy haguére ñe'ẽme. Hetaheta ojehesa'ỹijóvo hembiapo ko'ã seis áño ohasava'ekuépe hina ohechauka heta haihára ojetaky inglé rehe. Upéicharõ, mba'épiko ikatu ojapo umi investigadór oimo'ãramo ikatuha oñemboyke hembiapo inglé oguerekógui? Ikatu ningo oĩ noñeha'ãiva okomparti hembiapokue. Ambue katu ikatu opaga peteĩva oñemyatyrõ haĝua ikuatiahaipyre, upéva ojehepyme'ẽ heta pláta ha ogueraha are guivéma. Péva omoî chupekuéra ivaive chupekuéra ĝuarã ambuegui ndikatúigui oñemotenonde ambue tapicha ndive upéicharamo jepe ikua'aporã ha ikatupyryporã rehegua.

Oĩ tapicha oimo'ãva oiporúvo mba'e IAguigua oipytyvõva jehai ha ñembohasa ha'eha iporãva. Ha katu añeteháp, ohechauka oîha peteî ivaíva mba'éichapa ñamoherakuã invetigaciõ. Umi haihára ikatuva'erã oikuaauka hembiapo iñe'ẽitepe, natekotevẽi ombohasa. Avei, pe traduksiõ ndaha'éi perfekto opa mba'ere, ha hepyeterei.

Umi mba'e noñemoambuéiramo, ikatu oiko mokõi mba'e vai: hetavéva tapicha ndaikatumo'ãi okomparti hemiandu noñe'ẽporãigui inglés-pe, térã opavave oikuaava'erã inglés oike haĝuánte pe komunida IA -pe. Ni peteĩva ko'ãvagui ndaha'éi hekojojáva. Upéicharõ, mba'épa ikatu jajapo? Umi amandaje oîva'erâ tetã ñe'ême oîhápe, ha umi oñembohasava ombosako'íva ome'êva'erâ

Umi ohesa'ỹijóva ndohusgaiva'erã kuatiakuéra ñe'ẽ jejavy rehe. Upéva rangue ojehejava'erã haihárape oikuaauka hembiapokuéra opaichagua ñe'ẽme. Avei, umi oñe'ẽva inglé-pe oipytyvõva'erã opaga haguã traduksión, ndaha'éi umi tapicha noñe'ẽiva inglé-pe añónte. Tekotevẽ jajesareko opavave tapicha investigador-pe oike, taha'e ha'éva ñe'ẽ oñe'ẽva. Oñemokyre'ỹva'erã temimbo'e oñemohu'ãvape oikuaa haguã ambue ñe'ẽ avei techapyrârô ava ñe'ẽite, ajépa. Jaipotáramo peteĩ komunida investigación enteropaite tetã guigua añete, tekotevẽ jaguerohory ha ñamomba'eguasu opa ñe'ẽ, ndaha'éi inglé añónte

Ha otrave, ñamoĩva'erã ombo'évo hesakã porã umi compañero revisor-kuérape ani haguã he'i oikuaaha mba'e ñe'ẽ jeporupa oĩ ñe'ẽ "oĩporãha" umi kuatia ohesa'ỹijóva; upéva rangue, umi maranduhára ohesa'ỹijova'erã ambue tape oñemoherakuã haguã beca heta ñe'ẽme, ha'eháicha oikuave'ẽvo umi karai arandukuérape pa'ũ oikuaauka haguã hembiapo heta ñe'ẽme. Umi oñe'ẽva inglé avei opaga va'erã hikuái traducción rehegua, omoĩ rangue enteramente umi noñe'ẽiva inglés-pe. Umi kuatiahaipyre ha aty guasu omoĩva'erã viru servicio de traducción-pe ĝuarã-ndaha'éi inglé-pe añónte, ha katu inglé-gui. Ha umi marandu haihára momba'apohára, OpenReview-icha, oguerekoveva'erã avei oñembohasa haĝua ñe'ẽasa.

Ha'eveite, tekotevêterei pe mba'epohýi ñemoambue rehegua ani oî yvypóra oñemboykéva mba'éichapa oñe'ẽ ohaiháicha hikuái: mayma yvypóra ko omba'apóva hendivekuéra opaite yvy ape ári pe omba'apova'erã oñemoarandu haguã oaguanta ha oñemomba'e haguã ambueñe'ẽrehe. Añeteháp ñañeha'ãramo jaguereko peteĩnte komunida IA opaite yvy ape áriguigua, ñadesafiava'erã mba'éichapa inglés ocontrola mba'éichapa oñemombe'u IArehe-pe.

# वैश्विक एआई समुदाय को भाषा-विविध प्रकाशन की आवश्यकता है

**हेली लेप्प और पार्थ सरीन**

**वंदना सरीन और विवेक सरीन ने इस शोध का हिंदी में अनुवाद किया**

विश्व मेंAI का निर्माण कौन करता है? इस संसार पर तैनात किए जा रहे मॉडलों का अनुसंधान, डिजाइन, निर्माण, प्रशिक्षण, मूल्यांकन और विक्रय कौन करता है? अंतरराष्ट्रीय स्तर पर लोगों का विविध नेटवर्क AI की प्रौद्योगिकियों में योगदान देता है, जिसमें शामिल हैं अज्ञात मजदूर जो ऊर्जा और हार्डवेयर के लिए भौतिक संसाधनों को निकालते हैं [7], न्यून वैतनिक मजदूर जो लेबल और लेबल और लेबल करते हैं [5]; और अवैतनिक "उपयोगकर्ता" जिनके श्रम को AI कंपनियों ने सृजन के रूप में नहीं, बल्कि उपभोग के रूप में वर्गीकृत किया है [10]. AI शोधकर्ता सुर्खियों में हैं। आमतौर पर उच्च वेतन पाने वाले कॉर्पोरेट कर्मचारी या उच्च कमाई की क्षमता वाले विद्वान, AI शोधकर्ताओं के पास इस तकनीक के प्रक्षेपवक्र को निर्धारित करने में एक बड़ी आवाज है, ऐसी आवाज जिसे वे प्रकाशनों और सम्मेलन की कार्यवाही के माध्यम से पेश करते हैं। बार-बार, इस समुदाय से नस्ल, लिंग, देश और उम्र में विविधता की कमी के लिए पूछताछ की गई है। परिणामस्वरूप, आत्मीयता समूहों का प्रसार हुआ है; सम्मेलनों ने पूर्वाग्रह-विरोधी नीतियों को अपनाया है, और स्वयंसेवकों ने उन समूहों के लिए प्रशिक्षण के अवसरों का विस्तार करने के लिए अथक प्रयास किया है, जिन्होंने ऐतिहासिक रूप से कंप्यूटिंग समुदायों में भेदभाव का सामना किया है। फिर भी विविधता का एक प्रमुख घटक अनसुलझा है, जिसके बारे में हम तर्क देते हैं कि यह इस शक्तिशाली उद्योग की निष्कर्षण और असंतुलित प्रकृति को बनाए रखने में मदद करता है। इस वैश्विक समुदाय में एआई शोधकर्ता बनने के लिए अंग्रेजी में लिखना और बोलना आवश्यक है।

कंप्यूटर विज्ञान कि शीर्ष 100 रैंक वाली सम्मेलन कार्यवाही और पत्रिकाएँ अंग्रेजी में प्रकाशित होती हैं [1]। कई देशों में जहां अंग्रेजी प्राथमिक भाषा नहीं है, सम्मेलन की कार्यवाही आधिकारिक तौर पर अंग्रेजी में आयोजित की जाती है। यहां तक कि जो विद्वान अंग्रेजी सीखने में काफी समय और खर्च लगाते हैं, उन्हें भी सहकर्मी समीक्षा में अस्वीकृति का सामना करना पड़ सकता है। वास्तव में, ICLR की कार्यवाही के पिछले छह वर्षों की ऐतिहासिक समीक्षाओं से पता चलता है कि हजारों समीक्षाएँ लेखकों की भाषा की आलोचना करती हैं, स्पष्ट रूप से ("पेपर अंग्रेजी गलतियों से भरा है।") या परोक्ष रूप से ("हमें व्याकरण संबंधी त्रुटियां और खराब वाक्यांश वाले वाक्य मिले।") [12]। इस एकभाषी अनुसंधान उद्योग के प्रभाव व्यापक हैं: भारी असमान भाषाई उत्पादन और तथाकथित "कम-संसाधन" भाषाओं में मुद्दों पर कम ध्यान देने से लेकर [4, 11], वैश्विक शिक्षा और नियुक्ति में सीमाएं [8], और उन विद्वानों के लिए वास्तविक कर, जिन्हें कॉपी-संपादन के लिए भुगतान करना पड़ सकता है [2, 3]।

एक शोधकर्ता क्या करता है जब उसे लगता है कि अंग्रेजी के बारे में समीक्षक की धारणा के आधार पर उसके काम को अस्वीकार कर दिया जा सकता है? सबसे पहले, कुछ लोग अपना काम सबमिट नहीं करेंगे और अनुसंधान समुदाय में भाग नहीं लेंगे। दूसरा, कुछ लोग अपना काम प्रस्तुत करेंगे, लेकिन अपने लेखन का अनुवाद करने या उसे "पेशेवर" बनाने के लिए किसी को भुगतान करेंगे। यह प्रक्रिया महंगी है और इसके लिए अतिरिक्त समय की आवश्यकता होती है, जो इतनी तेजी से आगे बढ़ने वाले क्षेत्र में उन विद्वानों को प्रतिस्पर्धी नुकसान में डालती है।

स्वचालित लेखन-सहायता और अनुवाद सॉफ्टवेयर की बढ़ती उपलब्धता को समावेशन के लिए एक वरदान के रूप में मनाया गया है, जिससे विद्वान मोनोलिंगुअल प्रकाशन में सहकर्मी समीक्षा से पहले अपने लेखन का तुरंत अनुवाद कर सकते है। कई विद्वान प्रकाशन के लिए अपने लेखन को "सुधारने" करने के लिएChatGPT का उपयोग करते हैं। हम इस घटना को समावेशन के समाधान के रूप में नहीं, बल्कि भाषाई बहिष्करण के एक लक्षण के रूप में मानते हैं। लेखकों को अनुवाद की मध्यस्थता के बिना अपनी पसंद की भाषा में अपने निष्कर्षों का वर्णन करने का अधिकार होना चाहिए। भविष्य के अध्ययन के लिए प्रारंभिक साक्षात्कार में, बहुभाषीICLR विद्वानों ने संकेत दिया कि उन्हें लगता है कि जब वे खुद को एक भाषा से दूसरी भाषा में व्यक्त करते हैं तो उनका व्यक्तित्व अलग-अलग होता है। अनुवाद कभी भी एक-से-एक नहीं होता है। केवल एक भाषा में प्रकाशित करने से समुदाय प्रतिनिधित्व के अन्य तरीकों की विशाल विविधता खो देता है। इसके अलावा, इंग्लिश में अनुवाद करने से हम एक समावेशी वैश्विक समुदाय की ओर जाने के बजाय, उससे विमुख हो जायेंगे।

हस्तक्षेप के बिना, दो संभावनाएँ क्षितिज पर दिखती हैं। AI प्रकाशन समुदाय दुनिया के अधिकांश भाषा समूहों और लोगों का बहिष्करण जारी रखेगा, और दुनिया में केवल सीमित लोग जो दीर्घकालिक अंग्रेजी प्रशिक्षण प्राप्त करते हैं, वे भाग लेंगे। वैकल्पिक रूप से, वर्तमान प्रकाशन अवसंरचना केवल अंग्रेजी शिक्षा और प्रकाशन में वृद्धि की मांग करेगी। कंप्यूटिंग प्रशिक्षण को अंग्रेजी-भाषा प्रशिक्षण के साथ जोड़ा जाएगा, जिससे दुनिया भर में उच्च वेतन वाले रोजगार और प्रौद्योगिकी अनुसंधान में बौद्धिक योगदान के लिए अंग्रेजी को एक आवश्यकता के रूप में आगे बढ़ाया जाएगा। इनमें से कोई भी वैश्विक अनुसंधान समुदाय के लिए सहनीय भविष्य नहीं है। तो, हम यहाँ से कहाँ जा सकते हैं?

सबसे पहले, सम्मेलनों को उस देश की भाषा(भाषाओं) में प्रशासित किया जाना चाहिए जिसमें वे आयोजित किए जाते हैं। आयोजकों को अनुवाद सेवाओं को किराए पर लेना चाहिए, और मौजूदा कंप्यूटिंग समुदाय में अंग्रेजी और अन्य भाषाओं को बोलने वाले विद्वानों को अपनी अन्य भाषाओं में प्रस्तुत करने के लिए प्रोत्साहित करना चाहिए।

दूसरा, हमें सहकर्मी समीक्षकों को स्पष्ट निर्देश शामिल करना चाहिए कि वे जिन पत्रों की समीक्षा करते हैं उनकी भाषा "उपयुक्तता" पर निर्णय न लें [6]; इसके बजाय, संपादकों को बहुभाषी छात्रवृत्ति प्रकाशित करने के लिए अन्य विकल्प तलाशने चाहिए, जैसे कि विद्वानों को कई भाषाओं में अपना काम साझा करने का अवसर प्रदान करनाचाहिए। अंग्रेजी बोलने वालों को अनुवाद की लागत का बोझ पूरी तरह से अंग्रेजी नहीं बोलने वालों पर डालने के बजाय खुद वहन करना चाहिए। प्रकाशनों और सम्मेलनों को अनुवाद सेवाओं के लिए धन अलग रखना चाहिए - न केवल अंग्रेजी में, बल्कि अंग्रेजी से भी। प्रकाशन अवसंरचना, जैसे OpenReview, में अनुवाद प्रस्तुत करने के लिए स्थान भी शामिल होना चाहिए।

अंत में, यह जरूरी है कि बदलाव का बोझ उन लोगों पर न पड़े जो प्रकाशन में भाषाई रूप से हाशिए पर हैं: इस वैश्विक समुदाय में हर किसी को भाषाई विविधता को सहन करने और अपनाने की दिशा में काम करना चाहिए। स्नातक छात्रों को भाषा पाठ्यक्रम लेने के लिए प्रोत्साहित किया जाना चाहिए [9]। यदि हम वास्तव में एक वैश्विक AI समुदाय की आकांक्षा रखते हैं, तो हमें कंप्यूटिंग में अंग्रेजी के आधिपत्य को चुनौती देनी होगी।

# کۆمەڵگەی جیهانی ژیری دەستکرد (ئەی ئای) پێویستی بە بڵاوکردنەوەی هەمەچەشنی زمانییە

**هەیڵی لیپ**

**پارث سارین**

**وەرگێڕان: ڕۆشنا عمر عبدالرحمن**

بڵاوکردنەوەی بەرهەمەکانیان بۆ ئەوەی دەستبەجێ نووسینەکانیان وەربگێڕن پێش پێشکەشکردنیان بۆ پێداچوونەوەی کۆنفرانسەکان و بڵاوکراوەکان، زۆرێک لە زانایان چات جی پی تی بەکاردەهێنن بۆ "چاککردنەوەی" نووسینەکانیان پێش بڵاوکردنەوەیان. هەرچەندە ئێمە ئەم دیاردەیە وەک چارەسەرێک بۆ گشتگیری نابینین، بەڵکو وەک نیشانەی دوورخستنەوەی زمانەوانی سەیری دەکەین. نوسەر هەقی خۆیەتی ئەنجامی توێژینەوەی بەو زمانەی کە دەیەوێ باس بکات، بێ ئەوەی بە ناوەندی وەرگێڕاندا بڕوات. لە چاوپێکەوتنی سەرەتایی بۆ توێژینەوەیەکی داهاتوو، زانا فرەزمانەکانی کۆنفرانسی ئای سی ئێڵ ئار ئاماژە بەوە دەکەن کە هەست دەکەن کەسایەتی جیاوازیان هەیە کاتێ خۆیان بە زمانی جیاواز دەردەبڕن. وەرگێڕان هەرگیز یەک-بۆ-یەک نیە، بۆیە بە تەنها بڵاوکردنەوە بە یەک زمان، کۆمەڵگەکە هەمەجۆری و زۆری ڕێگای تری زانین کە دەکرێ بەدەرکەوێ، لەدەست دەچێ. وە هەروەها، بەرهەمهێنانی توێژینەوە بە یەک زمان، خوێنەرانی زمانەکانی دیکەش پشتگوێ دەخا. بە بەردەوام بوونی وەرگێڕان بۆ زمانی ئینگلیزی وا دەکا بوارەکەمان لەبری ئەوەی بەرەو کۆمەڵگەیەکی جیهانی گشتگیرمان ببا، بەڵکو دوورمان دەخاتەوە.

بەبێ خۆ تێهەڵقورتاندن، دوو ئەگەرمان لەبەردەستایە. یان کۆمەڵگەی بڵاوکردنەوەی کۆنفرانسی ئای سی ئێڵ ئار بەردەوام دەبێت لە دوورخستنەوەی زۆرینەی گروپەکانی زمانی جیهان و بەو هۆیەوە خەڵکەکانیشیان، وە تەنها ئەو کەسانەی جیهان کە ڕاهێنانی درێژخایەنی ئینگلیزی وەردەگرن دەتوانن بەشداری بکەن لەو کۆمەڵگایانە. یانیش ژێرخانی بڵاوکردنەوەی ئێستا داوادەکات بە زیادکردنی هەوڵدان بۆ ئەوەی پەروەردە و خوێندن و بڵاوکردنەوە بەس بە زمانی ئینگلیزی بێ. ڕاهێنانی کۆمپیوتەر بە ڕاهێنانی زمانی ئینگلیزی دەبەسترێتەوە، کە وا دەکا زمانی ئینگلیزی لە سەرانسەری جیهاندا ببێتە مەرج بۆ دامەزراندن و مووچەی بەرز و بەشداریکردن لە توێژینەوەی تەکنەلۆژیادا. هیچ کام لەمانە داهاتوویەکی وا نین بۆ گەشەی کۆمەڵگەی توێژینەوەی تەواوی جیهانی. کەواتە، ئێمە لێرەوە دەتوانین بۆ کوێ بچین؟

یەک، هەندێکیان کارەکانیان بڵاو ناکەنەوە و بەشداری ناکەن لە کۆمەڵگەی توێژەران. ڕێکخەرانی بڵاوکراوە و کۆنفرانسەکان پێویستە خزمەتگوزاری وەرگێڕان بەکرێ بگرن، و هانی ئەو زانایانەش بدەن کە بەزمانی ئینگلیزی و زمانەکانی تر قسە دەکەن تا بە زمانی دیکە بەرهەمیان پێشکەش بکەن، ئەمەش وادەکا هەمەجۆریی زمانی کۆمەڵگەی کۆمپیوتەریی ئێستا بەدەربکەوێت.

دووەم، پێویستە ڕێنمایی ڕوون بۆ ئەو کەسانەی کە پێداچوونەوە دەکەن دابنێین کە بڕیار لەسەر "گونجاوی" زمانی ئەو توێژینەوە نەدەن کە پێداچوونەوەی بۆ دەکەن ( [6])؛ لەبری ئەوە، سەرنوسەران دەبێ لە ڕێگای دیکە بکۆڵنەوە بۆ بڵاوکردنەوەی سکۆلەرشیپی فرە زمانی، بۆنمونە دەرفەت پێشکەش بکەن بە زانایان بۆ ناردنی کارەکانیان بە چەندین زمانی جیاواز. پێویستە ئینگلیز زمانەکان هەر بارگرانییەکی تێچووی وەرگێڕان لەسەر خۆیان هاوبەشی بکەن، لەبری ئەوەی بە تەواوی بخەنە سەر ئەو کەسانەی کە بە ئینگلیزی قسە ناکەن. چاپکراوەکان و کۆنفڕانسەکان دەبێ پارە بۆ خزمەتگوزارییەکانی وەرگێڕان تەرخان بکەن- نەک تەنها بۆ وەرگێڕان بۆ زمانی ئینگلیزی، بەڵکو لە زمانی ئینگلیزییەوە بۆ زمانەکانی تر. ژێرخانی بڵاوکردنەوەی وەک ئۆپن ڕیڤیو، پێویستە شوێنێک بۆ پێشکەشکردن و دانانی وەرگێڕانەکان لەخۆبگرێت.

لە کۆتاییدا، زۆر گرنگە کە بارگرانی گۆڕانکاری لەسەر ئەو کەسانە نەبێت کە لە ڕووی زمانەوانییەوە پەراوێزخراون لە بڵاوکردنەوەدا: هەموو کەسەکانی ئەم کۆمەڵگە جیهانییە پێویستە هەوڵبدەن بۆ ئەوەی بەرگەگرتن و وەرگرتنی هەمەچەشنی زمانەوانی لەخۆبگرن. پێویستە خوێندکاری باڵای زانکۆکان ئەگەر پێویستیش نەبێ، پێویستە هانبدرێن تا خولی زمان وەربگرن ([9]). ئێمە دەبێت تەحەدای باڵادەستی زمانی ئینگلیزی بکەین لە کۆمپیوتەر، ئەگەر بەڕاستی ئاوات دەخوازین بۆ کۆمەڵگەیەکی جیهانی ژیری دەستکرد.

کێ بەرنامەکانی ژیری دەستکرد (ئەی ئای) جیهانی بەرهەمدەهێنێ؟ کێ ئەرکی لێکۆڵینەوە و دیزاین و دروستکردن و فێرکردن و هەڵسەنگاندن و فرۆشتنی ئەو مۆدێلانە لە ئەستۆ دەگرێ کە لە هەموو سەرانسەری دونیا بەکاردەهێنرێ؟ تۆڕی کەسانێکی هەمەچەشنی نێودەوڵەتی بەشداری تەکنەلۆژیای ناسراو بە ژیری دەستکرد (ئەی ئای) دەکەن، ئەمەش کرێکاری نادیار "تارمایی" ەکانیش لەخۆدەگرێتەوە کە سەرچاوە مادییەکان بۆ وزە و هاردوێر دەردەهێنن ([7])، ئەوانەی کە داهاتێکی زۆر کەمیان هەیە بەرامبەر بە ناولێنان (لەیبڵکردن)ە بەردەوامەکانیان ([5])، وە ئەو "بەکارهێنەر"انەش کە موچە وەرناگرن هەرچەندە ماندووبوونیان بە بەکارهێنان دیاری کراوە نەک بە ئیش و کار لە لایەن کۆمپانیاکانی ژیری دەستکرد (ئەی ئای) ([10]). دەست خۆشی و چاکەش هەمووی بۆ توێژەری ژیری دەستکرد (ئەی ئای)ەکان دەگەڕێتەوە. بەشێوەیەکی گشتیش کارمەندانی کۆمپانیا مووچە بەرزەکان یان زانا داهات زۆرەکان، ئەوانەن کە ئاڕاستەی تەکنەلۆژیای ژیری دەستکرد (ئەی ئای) بە دەنگی بەرز دیاریدەکەن، دەنگێک کە لە ڕێگەی بڵاوکراوەکان و کۆنفڕانسەوە بە زۆر کەس دەگات. هەر بە بەردەوامی دووبارە و سێبارە، ئەو کۆمەڵگەیە لێپرسینەوەی لەگەڵ کراوە لەبەر کەمی هەمەڕەنگی لە لایەنی ڕەگەزی، نەتەوەیی، و تەمەنی. لە ئەنجامی ئەمەشدا، گروپە لێکچووەکان بەرەو زیاربوون دەچن؛ کۆنفرانسەکانیش سیاسەتی دژە لایەنگرییان گرتۆتەبەر، وە خۆبەخشەکان بە ماندوو نەناسانیەوە کاریان کردووە بۆ فراوانکردنی دەرفەتی ڕاهێنان بۆ ئەو گروپانەی کە پێشتر تووشی ڕووبەڕووی توندڕەوی بوونە لە کۆمەڵگاکانی کۆمپیوتەردا. بەڵام بەشێکی دیاری هەمەجۆرییەکە بە بێدەنگی ماوەتەوە، ئەوەی کە ئێمە ئاماژەی پێ دەکەین یارمەتیدەرە تا خراپ بەکارهێنان و ئەو ناهاوسەنگییەی کە لەم پیشەسازییە زلهێزانە هەیە بەردەوام بێت. بۆ ئەوەی ببیتە توێژەری ژیری دەستکرد (ئەی ئای) لەم کۆمەڵگا جیهانییە، تاک پێویستە بتوانێ بە زمانی ئینگلیزی قسە بکا و بنوسێ.

١٠٠ بەرزترین ڕیزبەندی بڵاوکراوەکانی کۆنفرانس و ڕۆژنامەکان بە زمانی ئینگلیزی چاپ کراون. هەرچەندە لە شوێنی جیا جیا بەڕێوەدەچن، لە زۆربەی ئەو وڵاتانەی کە زمانی سەرەکیشیان ئینگلیزی نیە، بڵاوکراوەکانی کۆنفرانسەکان هەر بەو زمانەیە ([1]). تەنانەت ئەو زانایانەی کە کات و خەرجی یەکی بەرچاوی خۆیان دەخەنە بەر فێربوونی ئینگلیزی ئەکادیمی دەکرێ توشی ڕەتبوونەوە بن کاتێ بڵاوکراوەکانیان پێداچوونەوەی بۆ دەکرێ. لە ڕاستیدا، بە سەیرکردنی پێداچوونەوەی ڕەخنەکانی شەش ساڵی ڕابردووی کۆنفرانسی ئای سی ئێڵ ئار دەبیبین کە لە هەزاران پێداچوونەوە بەدرێژایی ئەم ساڵانە ڕەخنە لە زمانی نووسەر دەگرن بە شێوەی ڕاستەوخۆ ("بڵاوکراوەکە پڕە لە هەڵەی ئینگلیزی.") یان بە شێوەیەی ناڕاستەوخۆ ("هەڵەی ڕێزمانی زۆر و ڕستەی دەستەواژەی ناڕاستی تێدایە.") ([12]). کاریگەریە خراپەکانی ئەم پیشەسازییە توێژینەوە تاک زمانە زۆر و فراوانە: لە نایەکسانییەکی زۆر لە نێوان بڵاوکراوەی زمانەکانی دیکە بۆ کەم گرینگیدان بە کێشەی زمانە "کەم سەرچاوەکان" ([4, 11])، تا بەربەستی لە بواری پەروەردە و دامەزراندنی کەسانی بیانی ([8])، تا باجێکی دیفاکتۆ بۆ ئەو زانایەی کە دەبێت پارە بدەن بۆ پێداچونەوەی زمانەوانی بەرهەمەکانیان پێش پێشکەشکردن ([2, 3]).

توێژەر چی دەکات کاتێک پێی وایە ڕەنگە کارەکەی ڕەت بکرێتەوە بە هۆی تێڕوانینەکانی ئەو کەسەی پێداچوونەوە دەکات بەرامبەر بە ئینگلیزی ئەکادیمی بڵاوکراوەکەی؟ یەک، هەندێکیان کارەکانیان بڵاو ناکەنەوە و بەشداری ناکەن لە کۆمەڵگەی توێژەران. دوو، هەندێ کارەکانیان بڵاو دەکەنەوە بەڵام پارە دەدەن بە کەسێک بۆ وەرگێڕان یان بۆ ئەوەی نوسینەکە ئەکادیمیتر بکرێ. ئەم پرۆسەیەش خەرجی زۆرە و کاتی زیاتری دەوێ، کە لەم بوارە خێرا و پڕ کێبڕێییە وا دەکات زیان بە زانایان بگات و هێواشتر بن لە توێژینەوە و بڵاوکردنەوەی بەرهەمەکانیان..

بەم شێوەیە، زیاد بەردەستبوونی بەرنامەکانی یاریدەی نووسین و وەرگێڕانی ئۆتۆماتیکی وەک بەخششێک بۆ گشتگیرکردن دەبینرێت. ئەمەش ڕێگە بەو زانایانە دەدا کە ڕەنگە بە هۆی تاک زمانی دوور بن لە

# Uma comunidade global de IA precisa de publicações em línguas diversas

**Haley Lepp and Parth Sarin**

**Gabriel Poesia traduziu para Português**

Quem produz uma Inteligência Artificial (IA) global? Quem pesquisa, projeta, constrói, treina, avalia e vende os modelos implantados em todo o mundo? Uma rede internacionalmente diversificada de atores contribui para as tecnologias conhecidas como IA, incluindo os trabalhadores "fantasmas" que extraem recursos materiais de energia e para fabricação de hardware [7], os que ganham salários mínimos para rotular, rotular e rotular dados [5]; e os "usuários" que não recebem salários, cujo trabalho é classificado pelas empresas de IA não como criação, mas como consumo [10]. Quem ocupa os holofotes são os pesquisadores de IA. Normalmente, pesquisadores de IA – ou funcionários corporativos altamente remunerados, ou acadêmicos com alto potencial salarial —, têm uma voz desproporcional na determinação da trajetória dessa tecnologia, que projetam por meio de publicações e anais de conferências. Essa comunidade tem sido questionada repetidamente por sua falta de diversidade de raça, gênero, nacionalidade e idade. Como resultado, grupos de afinidade se multiplicaram; conferências adotaram políticas anti-vieses e voluntários trabalharam incansavelmente para expandir as oportunidades de treinamento para grupos que historicamente enfrentaram discriminação nas comunidades de Computação. Entretanto, há um componente importante da diversidade que permanece não abordado e que, como aqui argumentamos, ajuda a sustentar a natureza extrativa e desequilibrada desta poderosa indústria. Para ser um pesquisador de IA nesta comunidade global, é necessário escrever e falar em inglês.

Todos os 100 principais anais e periódicos de Ciência da Computação são publicados em inglês [1]. Apesar de suas localizações em todo o mundo, em muitos países nos quais o inglês não é a principal língua falada, os anais das conferências são oficialmente publicados em inglês. Mesmo os estudiosos que investem tempo e despesas consideráveis para aprender a produzir inglês acadêmico podem enfrentar rejeição na revisão por pares. De fato, uma análise das revisões históricas dos últimos seis anos dos anais do ICLR demonstrou que milhares de revisões ao longo dos anos criticam o idioma dos autores explicitamente ("O artigo está cheio de erros em inglês.") ou implicitamente ("Há inúmeros erros gramaticais e frases mal formuladas.") [12]. As ramificações dessa indústria de pesquisa monolíngue são amplas: desde a produção linguística altamente desigual e pouca atenção a questões em idiomas chamados de "pobres em recursos" [4, 11], passando por limitações na educação e contratação global [8], até um imposto, na prática, para estudiosos que precisam contratar serviços de edição e revisão antes de submeter artigos [2, 3].

O que um pesquisador faz quando acredita que seu trabalho pode ser rejeitado com base nas percepções dos revisores sobre seu inglês acadêmico? Alguns optam por não submeter seu trabalho, e simplesmente não participar da comunidade de pesquisa. Segundo, alguns enviam artigos, mas pagam alguém para traduzir ou "profissionalizar" sua redação. Tal processo é caro e requer tempo extra, o que em um campo tão acelerado coloca esses estudiosos em uma desvantagem competitiva.

Assim, a crescente disponibilidade de software de assistência à escrita e tradução automática tem sido celebrada como uma vantagem para a inclusão, permitindo que estudiosos que de outra forma estariam excluídos da publicação monolíngue traduzam instantaneamente seus textos antes da submissão à revisão por pares. Muitos estudiosos usam o ChatGPT para "corrigir" seus textos para publicação. No entanto, consideramos esse fenômeno não como uma solução para a inclusão, mas sim um sintoma da exclusão linguística. Autores devem ter o direito de descrever suas descobertas no idioma de sua escolha, sem a mediação da tradução. Em entrevistas preliminares que realizamos para um estudo futuro, pesquisadores multilíngues do ICLR indicaram que sentem ter personalidades diferentes quando se expressam em um idioma ou outro. A tradução também nunca é um-para-um: então, ao publicar em apenas um idioma, a comunidade perde a vasta diversidade de outras formas de conhecimento que poderiam ser representadas. Além disso, produzir pesquisas em um único idioma afasta os leitores de outros idiomas. A tradução generalizada para o inglês afastará nosso campo de uma comunidade global inclusiva, em vez de se aproximar dela.

Sem intervenção, duas possibilidades surgem no horizonte. Em um cenário, a comunidade de publicação de IA continuará a excluir a maioria dos grupos linguísticos do mundo e, portanto, as pessoas, e apenas o grupo restrito de pessoas que recebem extensos anos de treinamento em inglês poderão participar. Em outra possibilidade, a infraestrutura de publicação atual exigirá uma maior assimilação da educação e publicação exclusivamente em inglês. O treinamento em computação será combinado com o treinamento em inglês, empurrando o inglês ao redor do mundo como um requisito para empregos bem remunerados e contribuição intelectual para a pesquisa em tecnologia. Nenhum desses futuros são toleráveis para uma comunidade de pesquisa de fato global. Para onde podemos ir, então?

Primeiro, as conferências devem ser administradas nos idiomas dos países em que são realizadas. Organizadores devem contratar serviços de tradução e incentivar os estudiosos que falam inglês e outros idiomas a apresentarem em seus outros idiomas, revelando a diversidade linguística da comunidade de computação existente.

Em segundo lugar, devemos incluir instruções explícitas aos revisores para que não julguem a "adequação" linguística dos artigos que revisam [6]. Em vez disso, os editores devem explorar outras opções para publicar trabalhos acadêmicos multilíngues, como oferecer aos estudiosos a oportunidade de compartilhar seu trabalho em vários idiomas. Falantes de inglês devem compartilhar qualquer ônus de custo de tradução, em vez de colocá-lo inteiramente sobre aqueles que não dominam o inglês. Publicações e conferências devem reservar fundos para serviços de tradução - não apenas para o inglês, mas também do inglês. A infraestrutura de publicação, como o OpenReview, também deve incluir espaço para submissões de traduções.

Por fim, é imperativo que o ônus da mudança não recaia sobre as pessoas linguisticamente marginalizadas na publicação científica: todos nesta comunidade global devem trabalhar para aprender a tolerar e abraçar a diversidade linguística. Alunos de pós-graduação devem ser encorajados, se não obrigados, a fazer cursos de idiomas [9]. Se realmente aspiramos ter uma comunidade global de IA, devemos desafiar a hegemonia do inglês na computação.

# Inclusividad lingüística para una verdadera comunidad global de IA.

**Haley Lepp and Parth Sarin**

**Traducido por Matías Grinberg**

¿Quiénes producen una IA global? ¿Quiénes investigan, diseñan, construyen, entrenan, evalúan y comercializan los modelos que se consumen a lo largo del planeta? Una red internacional y plurinacional contribuye a la tecnología conocida como IA, incluyendo a los trabajadores "fantasma" que extraen los recursos materiales necesarios para la energía y el hardware [7]; trabajadores mal remunerados que dedican interminables horas etiquetando datos uno tras otro[5]. Y los no remunerados "usuarios" cuya labor no fue considerada creación sino consumo por las empresas tecnológicas[10].

En el centro de la escena, los investigadores de IA. Generalmente empleados corporativos que perciben una generosa remuneración, o académicos con alto potencial de ingresos, estos investigadores gozan de una voz privilegiada al momento de influir en la trayectoria de esta tecnología, una voz que se proyecta a través de publicaciones y conferencias. Una y otra vez, se les reclamó la escasa diversidad étnica, de género, nacionalidad y edad. Como resultado, proliferaron nuevos grupos de afinidad; las conferencias implementaron políticas en contra de los sesgos, y voluntarios trabajaron incansablemente para expandir las posibilidades de formación para poblaciones que fueron históricamente discriminadas en las comunidades de computación. Sin embargo, un componente importante de la diversidad continúa sin ser resuelto. Uno que consideramos un importante factor sostenedor de la naturaleza extractiva y desbalanceada de esta poderosa industria. Para ser un investigador de modelos de IA en esta comunidad global, uno debe escribir y hablar en inglés.

Las 100 revistas y actas de congresos de ciencias de la computación mejor puntuadas se publican en inglés [1]. A pesar de encontrarse distribuidas por todo el planeta, en muchos lugares los congresos se hacen oficialmente en inglés a pesar de que no sea su idioma principal. Incluso investigadores que invierten una cantidad significativa de tiempo y recursos en la producción de inglés académico pueden enfrentarse a rechazos en la revisión por pares por razones directamente relacionadas al idioma. De hecho, el análisis de reseñas históricas a lo largo de los años critican el lenguaje de los autores tanto explícitamente ("Este artículo está lleno de errores de inglés") o implícitamente ("Hay numerosos errores gramaticales y oraciones mal formuladas")[12]. Las ramificaciones de esta industria monolingüe son extensas: desde outputs lingüísticos notablemente desiguales fuertemente desiguales y poca atención puesta en idiomas de "pocos recursos" [4, 11], a limitaciones en la educación global y contrataciones[8], y hasta impuestos de facto para académicos que deben contratar a editores de texto previamente a postular sus trabajos [2, 3].

¿Qué debe hacer un investigador cuando sospecha que su trabajo podría ser rechazado por juicios de los revisores relacionados al uso del inglés? Primero, algunos elegirán no postular su trabajo, simplemente dejando de participar en la comunidad científica. Segundo, algunos sí lo harán, pero no sin antes pagarle a alguien para traducir o "profesionalizar" su escritura. El costo en tiempo y dinero de este proceso implica para los investigadores una desventaja competitiva en un campo de rápida evolución.

Por esto, la creciente disponibilidad de software como asistente automático para la escritura y traducción fue celebrada como un hito para la inclusividad, permitiendo la traducción instantánea a académicos que si no quizás habrían sido excluidos. Muchos usan ChatGPT para "corregir" su escritura. Sin embargo, no lo consideramos como una solución para la inclusión, sino como un síntoma de la exclusión lingüística. Los autores deberían tener el derecho de presentar sus descubrimientos en el idioma de su elección, sin la mediación forzosa de una traducción. En entrevistas realizadas para un estudio por ser publicado, académicos políglotas reflejaron que sienten tener distintas personalidades según el idioma con el que se expresan. La traducción nunca es uno-a-uno, por lo que publicando solo en un idioma, la comunidad pierde la vasta diversidad en los modos de conocimiento posibles que podrían estar representados. Además, la producción de investigación en un solo idioma aliena a los lectores de otros idiomas. La traducción extendida y exclusiva del inglés no acercará a nuestra disciplina, sino que la alejará de una comunidad global inclusiva.

Sin intervención, dos posibilidades asoman por el horizonte. Las publicaciones de IA y su comunidad continuarán excluyendo a la mayoría de los grupos lingüísticos del mundo y a sus hablantes, y sólo quienes reciban una extensa educación en inglés participarán. Alternativamente, la infraestructura actual de publicación concentrará cada vez más en el inglés a la educación y a la investigación. La formación en computación irá de la mano con la formación en inglés, inculcándolo en todo el mundo como un requisito para lograr empleos bien remunerados y contribuciones intelectuales a la investigación en tecnología. Ninguno de estos son futuros admisibles para una comunidad de investigación verdaderamente global.

Para empezar, las conferencias deberían ser impartidas en el/los idioma(s) del país en donde sean realizadas. Los organizadores deberían contratar servicios de traducción, y fomentar a los participantes multilingües que en lugar de presentar en inglés elijan los suyos de preferencia, develando la diversidad lingüística de la comunidad de computación existente.

Segundo, deberíamos dar instrucciones explícitas a los revisores de no juzgar el nivel de "adecuación" del lenguaje en los artículos que evalúan[6]; en su lugar, los editores deberían explorar otras opciones para admitir múltiples idiomas, como ofrecer la posibilidad de publicar en varios idiomas. Las personas angloparlantes deberían compartir el costo de la traducción, en lugar de cargarla por completo en quienes no hablan inglés. Las publicaciones y conferencias deben asignar fondos para servicios de traducción, tanto al inglés como al inglés. La infraestructura para la publicación, como OpenReview, también debería incluir espacios para la presentación de traducciones.

Por último, es imperativo que la carga de estos cambios no recaiga en quienes son actualmente lingüísticamente marginalizados: la comunidad global completa debería estar trabajando en pos de aprender a tolerar y acoger la diversidad lingüística. Las instituciones educativas deberían fomentar, o incluso requerir, a estudiantes de grado el estudio de idiomas [9]. Si realmente aspiramos a tener una comunidad global de IA, debemos desafiar la hegemonía del inglés en la computación.